\documentclass[letterpaper]{article} 
\usepackage{aaai20}  
\usepackage{times}  
\usepackage{helvet} 
\usepackage{courier}  
\usepackage[hyphens]{url}  
\usepackage{graphicx} 
\usepackage{graphicx}  
\usepackage{url}
\usepackage{amsmath}
\usepackage[utf8]{inputenc}
\usepackage[english]{babel}
\usepackage{amsthm}
\usepackage{amsfonts}
\usepackage[titlenumbered,ruled,linesnumbered]{algorithm2e}
\usepackage{multirow}
\usepackage{caption}
\usepackage{subcaption}
\usepackage{bbm}
\nocopyright

\title{Identifying Hidden Buyers in Darknet Markets via Dirichlet Hawkes Process}
\author{Panpan Zheng,\textsuperscript{1} Shuhan Yuan,\textsuperscript{1} Xintao Wu,\textsuperscript{1} Yubao Wu\textsuperscript{2}  \\
\textsuperscript{1}{University of Arkansas, \{pzheng,sy005,xintaowu\}@uark.edu} \\
\textsuperscript{2}{Georgia State University, ywu28@gsu.edu} }

\begin{document}

\maketitle

\begin{abstract}
    The darknet markets are notorious black markets in cyberspace, which involve selling or brokering drugs, weapons, stolen credit cards, and other illicit goods. To combat illicit transactions in the cyberspace, it is important to analyze the behaviors of participants in darknet markets. Currently, many studies focus on studying the behavior of vendors. However, there is no much work on analyzing buyers. The key challenge is that the buyers are anonymized in darknet markets. For most of the darknet markets, We only observe the first and last digits of a buyer's ID, such as ``a**b''. To tackle this challenge, we propose a hidden buyer identification model, called \textit{UNMIX}, which can group the transactions from one hidden buyer into one cluster given a transaction sequence from an anonymized ID. UNMIX is able to model the temporal dynamics information as well as the product, comment, and vendor information associated with each transaction. As a result, the transactions with similar patterns in terms of time and content group together as the subsequence from one hidden buyer. Experiments on the data collected from three real-world darknet markets demonstrate the effectiveness of our approach measured by various clustering metrics. Case studies on real transaction sequences explicitly show that our approach can group transactions with similar patterns into the same clusters.
\end{abstract}

\section{Introduction}

Darknet markets are online commercial websites that strongly provide privacy guarantees to both vendors and buyers. The markets are hosted in the darknet based on TOR service to hide the IP address and adopt cryptocurrencies, such as Bitcoin, as payment methods. Due to its anonymity, most of the transactions on darknet markets are related with trading illicit goods, such as illicit drugs, stolen credit cards, or even weapons.

To combat illicit transactions in the cyberspace, it is important to analyze the behavior of participants in darknet markets. Currently, many studies focus on studying the behavior of vendors, such as linking multiple accounts from a same vendor (Sybil accounts) \cite{Tai:2019,Zhang:2019,Wang:2018:YYP:3196494.3196529}. However, there is no much work on analyzing buyers. One of the key challenges is that the buyers are anonymized in darknet markets.

\begin{figure}[!th]
    \centering
    \includegraphics[width=0.48\textwidth]{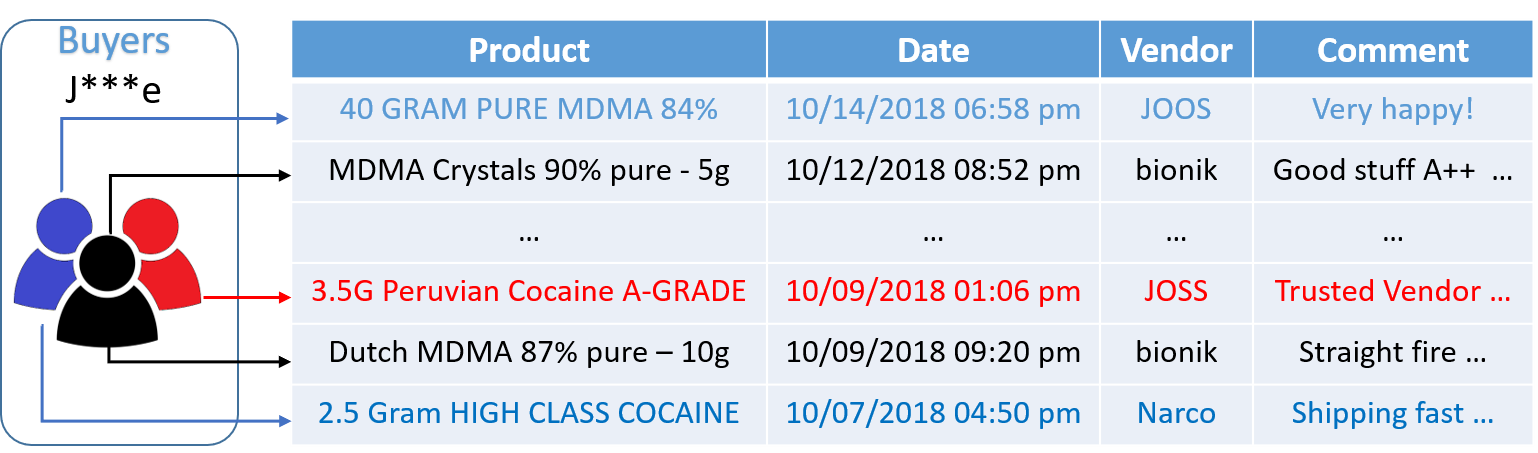}
    \caption{Illustrative example of transaction sequence from anonymized ID ``J***e'', where transactions are from various real-world buyers.} \label{fig:mixed}
\end{figure}

In order to protect the buyers' privacy and encourage the buyers to publish their comments, most of the darknet markets only reveal the first and last digits of a buyer's ID, such as ``a**b'', on the comment page. As a result, one observed anonymized ID can link to many different real-world buyers. Figure \ref{fig:mixed} shows an illustrative example of one mixed transaction sequence from the anonymized ID {\em J***e}. Each transaction contains information about four attributes: product, date, vendor, and comment in addition to the anonymized buyer name information. Our goal is to group those mixed transactions into clusters based on both content and temporal dynamics such that each cluster contains all transactions from one particular real-world user. Such disambiguation will allow us to learn transaction patterns of darknet markets and predict future transactions.

In this work, we propose \textit{UNMIX}, a hidden buyer identification model, based on the Dirichlet-Hawkes Process (DHP) \cite{du2015dirichlet}, which is able to group continuous-time transactions for each hidden buyer by modeling temporal dynamics, product, title, and comment of each transaction. Temporal dynamics here refers to the time patterns of purchase behavior in Darknet market. Each buyer has its own temporal dynamics. For instance, buyer ``Jaae'' often purchases one kind of heroines on Monday night once a week while buyer ``Jbbe'' buys the same drug on Tuesday and Thursday (twice a week). The output from our model are clusters, each of which contains transactions from one specific buyer. The idea of our proposed model is to have the Hawkes process model the intensity rate of transactions, while the Dirichlet process captures the buyer-transaction cluster relationships (i.e., each cluster contains transactions from one buyer). In practice, different but similar buyers may have similar transaction patterns and even the same user, his/her transaction patterns may change along with the time. However, in our darknet market scenario, each anonymized transaction sequence (e.g., with ID $J**e$) only consists of a few real buyers. Based on our observation, these few real buyers tend to have different transaction patterns and with rich information like transaction comment, time and vendor, our  UNMIX that groups transactions based on similar pattern can actually identify the hidden buyers.

UNMIX is a novel approach to achieve hidden buyer identification by integrating all the information associated with transactions, including temporal dynamics, products, comments, and vendors. The prior of each transaction belonging to one hidden buyer is determined by its temporal dynamics as different hidden buyers often exhibit different temporal dynamics. Specifically, the Hawkes process, one type of temporal point processes, is adopted to model the self-excitation phenomenon among transactions over continuous-time (e.g., buying illicit drugs in the past can raise the probability of buying them again in the future). The temporal dynamics of each identified hidden buyer is then characterized by one Hawkes process. Besides the temporal dynamics, the texts in product titles and comments and vendor involved in transactions are also incorporated into our model, which are characterized by a multinomial distribution and a categorical distribution, respectively. Meanwhile, by leveraging the Dirichlet process, the proposed model complexity grows as more transactions are collected over time, so our approach allows the number of hidden users from a mixed transaction stream that is not a-priori known or fixed.

The main contributions of our model are as follows. First, UNMIX does not need to assign a fixed number of hidden buyers underlying the unlimited number of transactions from one anonymized ID. Second, together with transaction content information, the temporal information provides important clues that improve accuracy in identifying hidden buyers in the same darknet markets. Third, experimental results on three real-world darknet markets indicate UNMIX is able to identify the various hidden buyers with different transaction patterns.

\section{Related Work}
\subsection{Darknet Market Analysis}
Darknet Markets are online markets hosted on the Tor service and guarantee strong anonymity property to participants. As a result, the darknet markets involve in illegal activities online. For the sake of public interests, the authorities and researchers have a growing interest to understand the darknet markets. Researchers have collected a large amount of data from darknet markets to analyze the active vendors, buyers, and goods being sold over time so that we can understand the growth of the darknet market ecosystem \cite{christin2013traveling,soska2015measuring}. \cite{dittus2018platform} conduct empirical studies to understand the supply chain underlying the markets.  Besides analyzing the volumes of whole darknet markets, some studies analyze specific categories in the darknet markets. For example, \cite{broseus2016studying} investigate the structure and organization of illicit drug trafficking.
Since the darknet markets have strong correlations with cybercrime, \cite{van2018plug} focus on measuring the commoditization of cybercrime via darknet markets.

Many researchers target on the micro-level analysis, which studies the participants in the darknet markets. Due to the anonymity of darknet markets, the challenge of analyzing the behavior of participants in the darknet market is how to link user identities in the markets.
Recently, several studies aim to link multiple accounts created by a real-world vendor \cite{Wang:2018:YYP:3196494.3196529,Tai:2019,Zhang:2019}. The key idea of these studies is based on ``stylometry'' analysis, which is originally used to attribute authorship to anonymous documents.  For example, \cite{Zhang:2019} link multiple vendors by analyzing the styles of the product pictures and descriptions published by vendors. Unlike matching vendors which can adopt lengthy product descriptions and photos, the information can be used for identifying hidden buyers is very limited. To the best of our knowledge, how to identify hidden buyers in the darknet markets has not been studied in the literature.

\subsection{Sequential Data Clustering}
Identifying hidden buyers from a mixed transaction sequence can be viewed as a task for clustering sequential data. The widely-used models for clustering data from the topic modeling literature are the Latent Dirichlet Allocation (LDA) \cite{blei2003latent}, where the number of topics is fixed, and its improved model, Hierarchical Dirichlet Process (HDP) \cite{teh2005sharing} with an unbounded number of topics. Many models are further proposed to fit the scenarios with online streaming text data \cite{Wang2008Continuous,ahmed2011online,Liang2016Dynamic}. Recently, several studies further incorporate temporal dynamics to group streaming data \cite{du2015dirichlet,mavroforakis2017modeling,xu2017dirichlet,seonwoo2018hierarchical}. For example, the Dirichlet Hawkes Process adopts the temporal point process, e.g., Hawkes process, to model the continuous-time information and the Dirichlet Process to solve the clustering problems \cite{du2015dirichlet}. In our work, we adapt the Dirichlet Hawkes Process for hidden buyer identification using the Hawkes process to model the temporal dynamics, the multinomial distribution to model texts in product titles and comments, and the categorical distribution to model vendors involved in transactions.

\section{Preliminary}

\subsection{Dirichlet Process}
The Dirichlet process (DP) is a Bayesian nonparametric model, which is parameterized by a concentration parameter $\alpha > 0$ and a base distribution $G_0$ over a space $\Theta$. It indicates that a random distribution $G$ drawn from DP is a distribution over $\Theta$, denoted as $G \sim DP(\alpha, G_0)$. The expectation of the distribution $G$ is the base distribution $G_0$. The concentration parameter $\alpha$ controls the variance of $G$ that a larger $\alpha$ leads to a tighter distribution around $G_0$.
DP is widely used for clustering with the unknown number of clusters.

The Dirichlet process can also be represented as the Chinese Restaurant Process (CRP). CRP assumes a restaurant with an infinite number of tables, and each of the tables can seat an infinite number of customers. Within the context of clustering, each table indicates a cluster while each customer is a data point. The simulation process of CRP is as follows:
\begin{enumerate}
    \item The first customer always sits at the first table.
    \item Customer $n$ ($n>1$) sits at:
        \begin{enumerate}
            \item a new table with probability $\frac{\alpha}{\alpha+n-1}$.
            \item an existing table $h$ with probability $\frac{n_h}{\alpha+n-1}$ where $n_h$ is the number of customers at table $h$.
        \end{enumerate}
\end{enumerate}
Let $\{\theta_1,...,\theta_n\}$ be a sequence sampled from $CRP$. The conditional distribution of $\theta_n$ can be written as:
\begin{equation}
\label{eq:dp}
    \theta_n|\theta_{1:n-1} \sim \frac{1}{\alpha+n-1} \big(\alpha G_0 + \sum_h n_h \delta_{\theta_h} \big),
\end{equation}
where $\delta_{\theta_h}$ is a point mass centred at $\theta_h$. Equation \ref{eq:dp} indicates that a new sample $\theta_n$ belongs to a new table with a constant probability or an existing table $h$ with probability proportional to $n_h$. A larger $n_h$ indicates a higher probability that a customer will belong to the table $h$. Hence, DP has a special clustering property that the rich gets richer.

\subsection{Temporal Point Process}
Temporal point process is a random process that models the observed random event patterns along the time. Given an event time sequence $\mathcal{T} =\{t_1, \cdots, t_n\}$, a temporal point process can be characterized by the \textit{conditional intensity function} which indicates the expected instantaneous rate of the next event at time $t$ ($t > t_n$):
\begin{equation}
\label{eq:lambda}
    \lambda^*(t)=\lambda(t|\mathcal{H}_{t_{n}})=\lim_{\text{d}t \rightarrow 0} \frac{\mathbb{E}[N([t,t+\text{d} t))|\mathcal{H}_{t_{n}}]}{\text{d}t},
\end{equation}
where $N([t,t+\text{d} t))$ indicates the number of events occurred in a time interval $\text{d} t$; $\mathcal{H}_{t_{n}}=\{t_{i} |t_{i}<=t_{n} \}$ is the collection of historical events until time $t_n$.

Let $f^*(t) = f(t|\mathcal{H}_{t_{n}})$ be the conditional density function of the event happening at time $t$ given the historical events up to time $t_{n}$, which is defined as
\begin{equation}
\label{eq:f}
    f^*(t)= \lambda^*(t) \cdot S^*(t) =\lambda^*(t) \cdot exp\Big(-\int_{t_{n}}^t \lambda^*(\tau)d\tau \Big),
\end{equation}
where $S^*(t)=S(t|\mathcal{H}_{t_{n}})=exp(-\int_{t_{n}}^t \lambda^*(\tau)d\tau)$ is the \textit{survival function} that indicates the probability that no new event has ever happened up to time $t$ since $t_n$.

With an observation window $[0, T]$, the joint likelihood of the observed sequence $\mathcal{T}$ is formalized as
\begin{equation}
\label{eq:tpp_likelihood}
    \mathcal{L}=\prod_{t_i \in \mathcal{T}} f^*(t_i) =\prod_{t_i \in \mathcal{T}} \lambda^*(t_i) \cdot exp\Big(-\int_{0}^{T} \lambda^*(\tau)d\tau \Big).
\end{equation}

{\bf\noindent Hawkes process.}
A Hawkes process is one type of temporal point process, which captures the self-excitation phenomenon among events \cite{hawkes1971spectra}. In the Hawkes process, the conditional intensity function is defined as:
\begin{equation}
    \lambda^*(t) = \lambda_0 + \sum_{t_i \in \mathcal{T}} \gamma(t,t_i),
\end{equation}
where $\lambda_0>0$ is the base intensity that indicates the intensity of events triggered by external signals instead of previous events; $\gamma(t,t_i)$ is the triggering kernel that is usually a monotonically decreasing function which ensures the recent events have higher influences on the intensity of next event. The Hawkes process models the self-excitation phenomenon that a new event arrival increases the conditional intensity of the oncoming event immediately and then decreases back towards $\lambda_0$ in the long term. Recently, the Hawkes process is widely used to model event patterns which are clustered, such as the information diffusion on social networks or the earthquake occurrences \cite{Zhao2015Seismic,Reinhart2017Review,Farajtabar2018Point}.

\section{Hidden Buyer Identification}
In a darknet market, a buyer purchases products from vendors , and then publishes comments about the products. Especially,it is noticed that we can't see the \textit{real user names} of buyers. Instead, what we can observe are some anonymized IDs, each of which contains an unbounded number of real buyers. Given a series of transactions marked by one specific anonymized ID, our goal is to uncover these real buyers, and then, based on them, to group the transactions. In our scenario, these \textit{distinctive real buyers} are named as \textit{hidden buyers}. Given a series of transactions $\mathcal{S}=\{e_1,...,e_n\}$ underlying one specific anonymized ID, its corresponding sequence of real buyers is denoted as $\mathcal{U}=\{u_1,...,u_n\}$ with one set of real buyers as $\{u_i\}$. Then, the \textit{hidden buyer} associated with one certain event $e$ is expressed as $u^* \in \{u_i\}$.

Formally, transaction $e$ in $\mathcal{S}$ is denoted as $e:=(t, u, v, p, c)$, which means that at time $t$, a buyer $u$ purchases a product $p$ from a vendor $v \in \mathcal{V}$, where $\mathcal{V} = \{v_1,...,v_n\}$ is the corresponding vendor sequence, and publishes a comment $c$. Since product titles and comments are both text information,  so we further combine them as a \textit{content} vector $\mathbf{w}$ by a bag of word model. Finally, we define one transaction in $\mathcal{S}$ as $e:=(t, u, v, \mathbf{w})$.
Note that since we only observe the time to publish a comment, in our scenario, we assume the operations, purchasing a product and publishing a comment, are synchronous.

To identify the \textit{hidden buyers}, we assume that different hidden buyers have their own unique \textit{hidden transaction patterns}. For example, buyer $\mathcal{A}$ always buys fentanyl from one certain vendor without comments, while buyer $\mathcal{B}$ often takes fentanyl from the same vendor as well but likes to leave the comments. Given this toy example, we are wondering if transactions with a similar purchasing pattern are associated with the same hidden buyer. To further explore and solve this problem, in this work, we aim to \textit{uncover} the \textit{mixed} transactions sequence marked by one anonymized ID ,and for this goal, we propose a novel identification framework named as \textit{UNMIX}.

UNMIX is a Dirichlet process framework with Chinese restaurant process as implementation. In UNMIX, each table encapsulates a marked Hawkes process model, which is for time and type information, and a bag-of-words model, which is for textual comment information. Here, each table corresponds to a real hidden buyer in our scenario. For one specific transaction, its hidden buyer assignment is based on a discrete probability distribution that is derived by posterior predictive distribution. The estimated occurrence likelihoods are related to the historical transactions from these hidden buyers. Hence, transactions with the similar patterns are easily going to the same hidden buyer and an oncoming transaction tends to be assigned to a hidden buyer (table) in which the majority of previous transactions (restaurant customer) are similar to it.

\subsection{Modeling Buyer Transactions}
From the perspective of features, we consider three categories of information: time, content (product titles and comments) and vendor. Each of them has its own distinctive characteristics and should be captured by different models. For instance, due to the drug addiction effects, once a user starts to purchase illicit drugs, he may keep purchasing constantly in a short period of time. Since the behavior of purchasing drugs is self-exciting, it is natural to adopt the Hawkes process to model the purchasing behavior in terms of time. Meanwhile, vendor type and content information are characterized by categorical and multinomial distributions, respectively. Given the unbounded number of hidden buyers in a dynamic transaction sequence, we adopt the Dirichlet process as a prior probability distribution to model the generation of hidden buyers.

Generally, UNMIX is a hierarchical framework with two layers: in the outer layer, it employs Dirichlet process to capture the diversity of \textit{hidden transaction patterns} for distinctive \textit{hidden buyers}; in the inner layer (inside the hidden buyers), it takes use of Hawkes process, multinomial distribution and categorical distribution to model the time, content and vendor type information, respectively.

\textbf{Intensity of the buyer transaction activity.}
We adopt the Hawkes process to model the buyer transactions over time. In our scenario, the sequence of transactions with the same anonymized ID are actually conducted by different hidden buyers. For each hidden buyer, we adopt one Hawkes process to model its temporal information. As a result, the intensity function of Hawkes process over the whole transaction sequence from all of existed hidden buyers is defined as:
\begin{equation}
\label{eq:hawkes}
    \lambda(t) = \lambda_0 + \sum_{h=1}^{H} \lambda_h(t),
\end{equation}
where $H$ is the total number of identified hidden buyers until time $t$. $\lambda_h(t)$ is the intensity of one certain hidden buyer $h$ and it can be expressed as follow:

\begin{equation}
\label{eq:hawkes_h_t}
    \lambda_h(t) = \sum_{t_i \in \mathcal{T}} \gamma_{h}(t,t_i)\mathbbm{1}[u_i=u^*_h],
\end{equation}
where $\mathcal{T} = \{t_1, t_2, \dots, t_n\}$ is the corresponding event time sequence of $\mathcal{S}$; $\gamma_{h}(t,t_i)$is the triggering kernel associated with one hidden buyer $u^*_h$; $u_i$ is the index of hidden buyer associated with the $i$-th transaction, and $\mathbbm{1}[u_i=u^*_h]$ denotes the $i$-th transaction has been assigned to the $h$-th buyer in Chinese restaurant process. Here, the triggering kernel function with $K$ base kernel functions is in the form as $\gamma_h(t,t_i)=\sum_{l=1}^K \alpha_h^l \kappa(\pi_l, t-t_i)$, where $\alpha_h^l > 0$ controls the self-excitation of the Hawkes process with $\sum_l \alpha_h^l = 1$, and $\pi_l$ is typical reference time point that controls the event decay. We adopt the Gaussian RBF kernel as the base kernel function.

\textbf{Distribution of content information (product titles and comments).}
Since both product titles and comments are text information, we represent them as a bag-of-word language model. We call both the product title and comment in a transaction as the \textit{content} of the transaction. As a result, we use a vector $\mathbf{w}_i$ to represent the content in transaction $e_i$, where each dimension refers to the frequency of the corresponding word sampled from a vocabulary $\mathcal{W}$. In particular, $\mathbf{w}_i$ provided by hidden buyer $h$, is describe as follow
\begin{equation}
\label{eq:multi}
    \mathbf{w}_i \sim Multi(\theta_h),
\end{equation}
where $\theta_h$ is the prior of multinomial distribution with size $|\mathcal{W}|$, which indicates the occurrence likelihood of each word in the content given the hidden buyer $u^*_h$.

\textbf{Distribution of vendors.}
In this work, we use vendor ID to indicate each vendor. Due to its dicreteness property, at each time $t_i$, the vendor type is sampled from a categorical distribution with the sample space size as $|\mathcal{V}|$:
\begin{equation}
\label{eq:cat}
    v_i \sim Cat(\eta_h),
\end{equation}
where $\eta_h$ is the prior of categorical distribution with size $|\mathcal{V}|$, which refers to the occurrence probability of each vendor type given the hidden buyer $u^*_h$.

\begin{figure}
  \centering
    \includegraphics[width=0.4\textwidth]{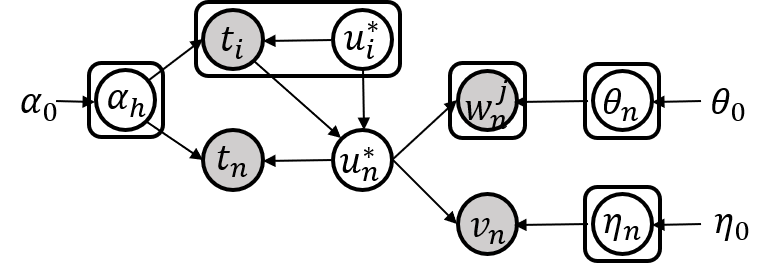}
\caption{Graphical representation of UNMIX}
\label{fig:dhp}
\end{figure}

\subsection{The Generative Process}

We can describe our model as a generative process similar to the CRP. At time $t$, the oncoming transaction $e$ may be  from either a new buyer or an existing buyer. To give a proper hidden buyer assignment of event $e$, our proposed framework UNMIX, which is running on a Dirichlet process, will dynamically reuse an existing \textit{hidden buyer} or generate a new one to adapt the upcoming event $e$. Concretely, hidden buyer $u$ of the oncoming event can be chosen in a metropolis sampling-based way
\begin{equation}
\label{eq:lambda_prior}
    u =
        \begin{cases}
        u^*_{H+1} & \text{with probability } \frac{\lambda_0}{\lambda(t)}  \\
        u^*_h & \text{with probability } \frac{\lambda_{h}(t)}{\lambda(t)},
        \end{cases}
\end{equation}
where $H$ is the number of existing \textit{hidden buyers} up to but not including time $t$; $\lambda_{h}(t)$ indicates the intensity of a Hawkes process for the hidden buyer $u^*_h$ defined in Equation \ref{eq:hawkes_h_t}. We can notice that $\lambda_0$ plays the similar role as the concentration parameter $\alpha$ in DP and the probability of $u$ belonging to $u^*_h$ is proportional to the intensity function $\lambda_h(t)$ from a Hawkes process.

The algorithm of the generative process is shown in Algorithm \ref{algr:dhp}, where $\lambda_0$ is the base intensity, $\alpha_0$ is the initial parameter setting of trigger kernels in Equation \ref{eq:hawkes_h_t}, $\eta_0$ and $\theta_0$ are the initial prior for the categorical and multinomial distributions. Line \ref{eq:sampling_time} samples the time $t$ via a Hawkes process. Based on temporal dynamics of historical events, line \ref{eq:sampling_hidden_buyer} chooses a proper hidden buyer for the current event at time $t$. Line \ref{eq:parameter_updating} mainly shows the updating of $\eta$s and $\theta$s to sample vendor type and content information for the current event in the next step. Given the priors ($\eta$s and $\theta$s) above, line \ref{eq:sampling_content} and \ref{eq:sampling_vendor_type} illustrate how to draw the corresponding \textit{content} and \textit{vendor type} information.

\begin{algorithm}
\SetAlgoLined
\SetKwInOut{Input}{Input}
\SetKwInOut{Output}{Output}
\Input{$\lambda_0$,$\alpha_0$, $\theta_0$, $\eta_0$}
\Output{$\{e_i:=(t_i, u_i, v_i, \mathbf{w}_i)\}_{i=1}^N$ where $N$ is the total number of transactions produced by the generative process algorithm.}
\For{$i=1,...,N$}{
	\label{eq:sampling_time}
    Sample the time $t_i \sim Hawkes(\lambda^*(t_i))$ \;
    \label{eq:sampling_hidden_buyer}
    Sample the hidden buyer $u_i$ for the transaction at time $t_i$ by Eq. \ref{eq:lambda_prior} \;
    \label{eq:parameter_updating}
    \uIf{$u_i == u^*_h$}{
       Reuse $\eta_h$ and $\theta_h$ for $\eta_i$ and $\theta_i$ \;
       }
    \uElse{
       Sample $\eta_i$ from $Dir(\eta|\eta_0)$, $\theta_i$ from $Dir(\theta|\theta_0)$, and $\alpha_i$ from $Dir(\alpha|\alpha_0)$ for the new user \;
      }
    \label{eq:sampling_content}
    Sample each word $\mathbf{w}_i$ in the content of transaction $e_i$ by Eq. \ref{eq:multi} \;
    \label{eq:sampling_vendor_type}
    Sample the vendor $v_i$ of transaction $e_i$ by Eq. \ref{eq:cat} \;
}
\caption{The generative process of UNMIX}
\label{algr:dhp}
\end{algorithm}

\subsection{Inference}
Given a sequence of transactions $\mathcal{S}=\{e_1,...,e_{n-1}\}$ from an anonymized ID, we aim to infer the hidden buyer (hidden transaction patterns) $u^*_h$ of the oncoming transaction $e_n$. We adopt a Sequential Monte Carlo (SMC) algorithm to sample the hidden buyer associated with each transaction $e_n$. SMC adopts a set of particles to approximate the posterior distribution $P(u_{1:n}|t_{1:n},\mathbf{w}_{1:n},v_{1:n})$, in which $P(u_{n}|u_{n-1},t_{1:n},\mathbf{w}_{1:n},v_{1:n})$ is taken as the proposal distribution. In particular, based on Figure \ref{fig:dhp}, the posterior distribution at time $t_n$ can be factorized as
\begin{align}
\label{eq:p_u}
\begin{split}
    &P(u_n|t_{1:n},\mathbf{w}_{1:n},v_{1:n}) \sim \\
    &P(v_n|u_n, rest) \cdot P(\mathbf{w}_n|u_n, rest)  \cdot P(u_n|t_n, rest),
\end{split}
\end{align}

In Equation \ref{eq:p_u}, the prior $P(u_n|t_n,rest)$ is  given by:
\begin{equation}
    P(u_n|t_n, rest) =
    \begin{cases}
        \frac{\lambda_0}{\lambda(t)} & \text{for new buyer}\\
        \frac{\lambda_{h}(t)}{\lambda(t)} & \text{for observed buyer $u_h$}
    \end{cases}
\end{equation}
where $\lambda_{h}(t):= \sum_{t_i \in \mathcal{T}} \gamma_{h}(t,t_i) \mathbbm{1}[u_i=u^*_h]$ indicates the intensity from buyer $u^*_h$. For the inference of $\alpha$, which is used to parameterize
the triggering kernels in the intensity function, we follow the literature \cite{4266870,10.2307/41058999} and update $\alpha$ by maximum likelihood estimation (Equation \ref{eq:tpp_likelihood}).

Based on the conjugate relation between the multinomial and Dirichlet distributions, the likelihood of the content distribution $P(\mathbf{w}_n|u_n, rest)$ is:
\begin{equation}
\begin{split}
    & P(\mathbf{w}_n|u_n, rest) \\
    & = \frac{\Gamma(C^{w_n} + 1)}{\prod_w^{\mathcal{W}} \Gamma(C^{w_n}_w + 1)} \cdot \frac{\Gamma(C^{u_n  \backslash w_n}+\sum_{w}^{\mathcal{W}} \theta^w_0)}{\prod_w^{\mathcal{W}}\Gamma(C^{u_n \backslash w_n}_w + \theta^w_0)} \cdot \\
    &  \frac{ \prod_w^{\mathcal{W}}\Gamma(C^{u_n \backslash w_n}_w + C^{w_n}_w + \theta^w_0)}{\Gamma(C^{u_n \backslash w_n} + C^{w_n} + \sum_{w}^{\mathcal{W}} \theta^w_0)},
\end{split}
\end{equation}
where $C^{u_n  \backslash w_n}$ and $C^{u_n \backslash w_n}_w$ indicate the total word count and the count of word $w$ appeared in the content from buyer $u_n$ excluding $w_n$, respectively; $C^{w_n}$ and $C^{w_n}_w$ refer to the total word count and the count of word $w$ in content $w_n$, respectively; $\theta^w_0$ is the value in Dirichlet prior for word $w$.

Similarly, the likelihood of the vendor distribution $P(v_n|u_n, rest)$ is:
\begin{equation}
    P(v_n = v|u_n, rest) = \frac{C^{u_n \backslash v_n}_{v} + \eta^v_0}{C^{u_n \backslash v_n}+ \sum_{v'}^{\mathcal{V}} \eta^{v'}_0},
\end{equation}
where $C^{u_n \backslash v_n}_{v}$ is the count of the vendor type $v$ from unique buyer $u_n$ excluding the current vendor $v_n$; $C^{u_n \backslash v_n}$ is the total number of vendors associated with the buyer $u_n$ excluding the current vendor $v_n$; $\eta^v_0$ is the value in Dirichlet prior for vendor $v$.

\section{Experiments}
\subsection{Datasets and Baselines.}
\begin{table}[]
\centering
\caption{Statistics of three darknet markets}
\label{tb:dataset}
\resizebox{.45\textwidth}{!}{
\begin{tabular}{|c|c|c|c|}
\hline
Darknet Markets             & Vendors & Anonymized Buyer IDs & Transactions \\ \hline
Wall Street Market         & 440     & 1896                 & 18603         \\ \hline
Empire Market       & 273     & 1492                 & 12937         \\ \hline
Dream Market        & 606     & 2587                 & 102378       \\ \hline
\end{tabular}
}
\end{table}

\begin{figure}
    \centering
    \includegraphics[width=.3\textwidth]{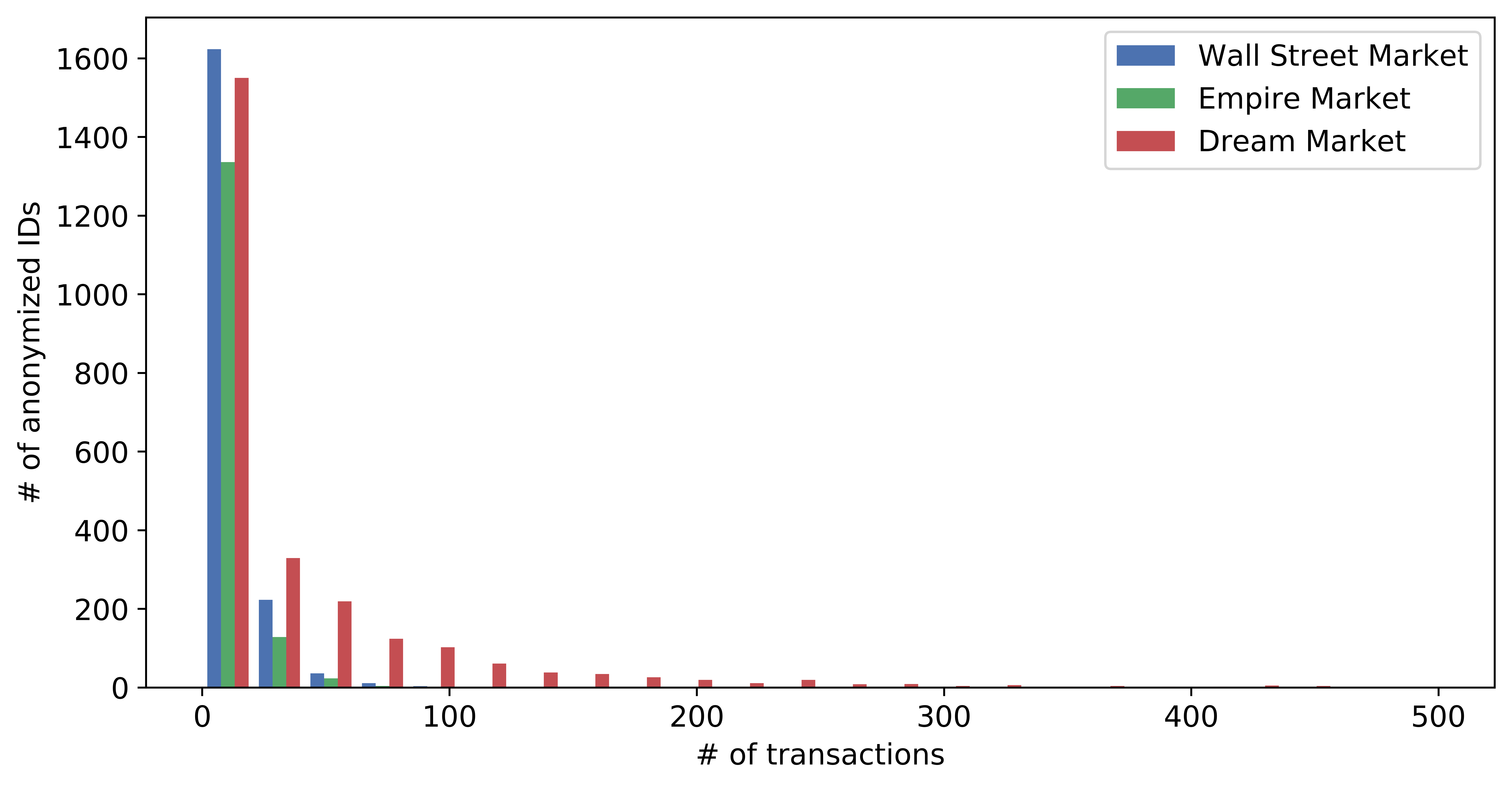}
    \caption{Distributions of transaction numbers conducted by anonymized IDs over three darknet markets}
    \label{fig:hist}
\end{figure}

{\bf \noindent Datasets.}
To evaluate our approach, we have crawled the data from three popular darknet markets, i.e., \textit{Dream Market}, \textit{Wall Street Market}, and \textit{Empire Market}. The statistics of the crawled darknet markets are shown in Table \ref{tb:dataset}. Figure \ref{fig:hist} further shows the distributions of transaction numbers over anonymized IDs in three darknet markets. Overall, it is a long-tail distribution, which indicates most of the anonymized IDs only conduct a small number of transactions.

Note that in the Dream Market, the buyers comment to the vendors instead of products. Hence, for the Dream Market, we only adopt the texts from comments as the content information.

{\bf \noindent Baselines.}
We compare our approach with two baselines.
\begin{itemize}
    \item Hierarchical Dirichlet Process (\textbf{HDP}) is a nonparametric Bayesian approach for topic modeling \cite{teh2005sharing}. We adopt DBSCAN to group the transactions, each of which is represented as the corresponding topic distribution. HDP only considers the information of product titles and buyer comments.
    \item Dirichlet Hawkes Process (\textbf{DHP}) is a simplified version of our approach which does not adopt the vendor information for clustering.
\end{itemize}

\subsection{Experiments on Transaction Sequences with Ground-truth}
{\bf \noindent Experimental setup.}
Due to the anonymity of darknet markets, it is infeasible to get the ground-truth regarding the actual buyers with the same anonymized id. To quantify the performance of our proposed approach, we propose a  procedure to generate transaction sequences with ground-truth. Specifically, based on our observations, the transactions conducted by one anonymized ID from one vendor in a short time have a high chance to be from one real-world buyer due to the consistent transaction behavior.

Therefore, for each darknet market, we first select $H$ anonymized IDs, where each anonymized ID has around five to eight transactions from one vendor in a month. Then, we combine all the transactions from these $H$ anonymized IDs to compose one transaction sequence and sort the sequence by transaction time. Hence, in this setting, we generate one transaction sequence for each darknet market, while each  transaction sequence is actual from various anonymized IDs. The goal of this task is to group transactions from one anonymized ID into one cluster. The statistics of transaction sequences with ground-truth are shown in Table \ref{tb:syn}.

We evaluate the performance by four clustering metrics, including \textit{adjusted rand score (ARS)}, \textit{normalized mutual information score (NMI)}, and \textit{V-measure score (V-score)}, \textit{homogeneity score (H-score)}. These metrics are computed by comparing with the ground-truth labels.

\begin{table}[]
\caption{Statistics of sequences with ground-truth}
\label{tb:syn}
\resizebox{.47\textwidth}{!}{
\begin{tabular}{|c|c|c|c|}
\hline
                           & Wall Street Market    & Empire Market   & Dream Market   \\ \hline
sequence length            & 42             & 188             & 229  \\ \hline
\# of anonymized IDs ($H$) & 6              & 27              & 36   \\ \hline
\end{tabular}
}
\end{table}

{\bf \noindent Experimental results.}
Table \ref{tb:cluster_syn} shows the clustering results on various transaction sequences. Overall, with incorporating the content, vendor, and time information for hidden buyer identification, our proposed approach achieves the good performance in terms of various clustering metrics. The performances of two baselines are worse than our proposed approach, which indicates without using vendor or temporal dynamics information could damage the performance of hidden buyer identification. Meanwhile, we can observe that the performance of the three approaches is reduced when the sequences become complicated. For example, our approach achieves the highest scores in Wall Street Market and the lowest scores in Dream Market. First, this is because the sequence of Wall Street Market is simple, which only consists of sequences from 6 anonymized IDs, while the sequence of Dream Market consists of 36 anonymized IDs. Moreover, for Dream Market, we only observe the comments as content information. Without using texts in product titles could damage the performance of clustering.

\begin{table}[]
\caption{Results of hidden buyer identification on transaction sequences with ground-truth}
\label{tb:cluster_syn}
\resizebox{0.48\textwidth}{!}{%
\begin{tabular}{|c|c|c|c|c|c|c|}
\hline
                              & Approaches   & ARS    & NMI    & V-score & H-score & \begin{tabular}[c]{@{}c@{}}\# of \\ IDs ($\hat{H}$)\end{tabular} \\ \hline
\multirow{3}{*}{\begin{tabular}[c]{@{}c@{}}Wall Street \\ Market\end{tabular}}  & HDP          & 0.1612 & 0.3380 & 0.3316  & 0.2777  & 3              \\ \cline{2-7}
                              & DHP          & 0.9675 & 0.9804 & 0.9802  & 0.9999  & 7              \\ \cline{2-7}
                              & Our approach & 0.9385 & 0.9627 & 0.9621  & 1.000   & 8              \\ \hline
\multirow{3}{*}{Empire Market}& HDP          & 0.0422 & 0.2537 & 0.2197  & 0.1464  & 7              \\ \cline{2-7}
                              & DHP          & 0.4874 & 0.8282 & 0.8281  & 0.8192  & 41             \\ \cline{2-7}
                              & Our approach & 0.5236 & 0.8549 & 0.8549  & 0.8588  & 44             \\ \hline
\multirow{3}{*}{Dream Market} & HDP          & 0.0215 & 0.3127 & 0.2773  & 0.1896  & 10             \\ \cline{2-7}
                              & DHP          & 0.1391 & 0.6171 & 0.6151  & 0.5697  & 45             \\ \cline{2-7}
                              & Our approach & 0.1831 & 0.6881 & 0.6878  & 0.6707  & 59             \\ \hline
\end{tabular}
}
\end{table}

We can notice that for Wall Street Market, DHP achieves a slightly better performance than our proposed approach. This is because the number of hidden buyers identified by DHP is close to the ground truth number. However, we argue that although we combine the sequence from different anonymized IDs to compose the sequence with ground truth, such sequence is only weakly-labeled since the short sequence from one anonymized ID could be actually from various hidden buyers. Based on our observation, our approach groups the subsequence from one anonymized ID into three clusters. However, these three hidden buyers do not share any common words in product titles and comments, which indicates the identified hidden buyers have different patterns such that they buy different products and have different comment styles. The identified three hidden buyers based on our approach have high chance to be three different real world buyers from the content aspect. Hence, the identification result of our approach on Wall Street Market is also reasonable.

\begin{figure}
    \centering
     \begin{subfigure}[b]{0.23\textwidth}
         \centering
         \includegraphics[width=\textwidth]{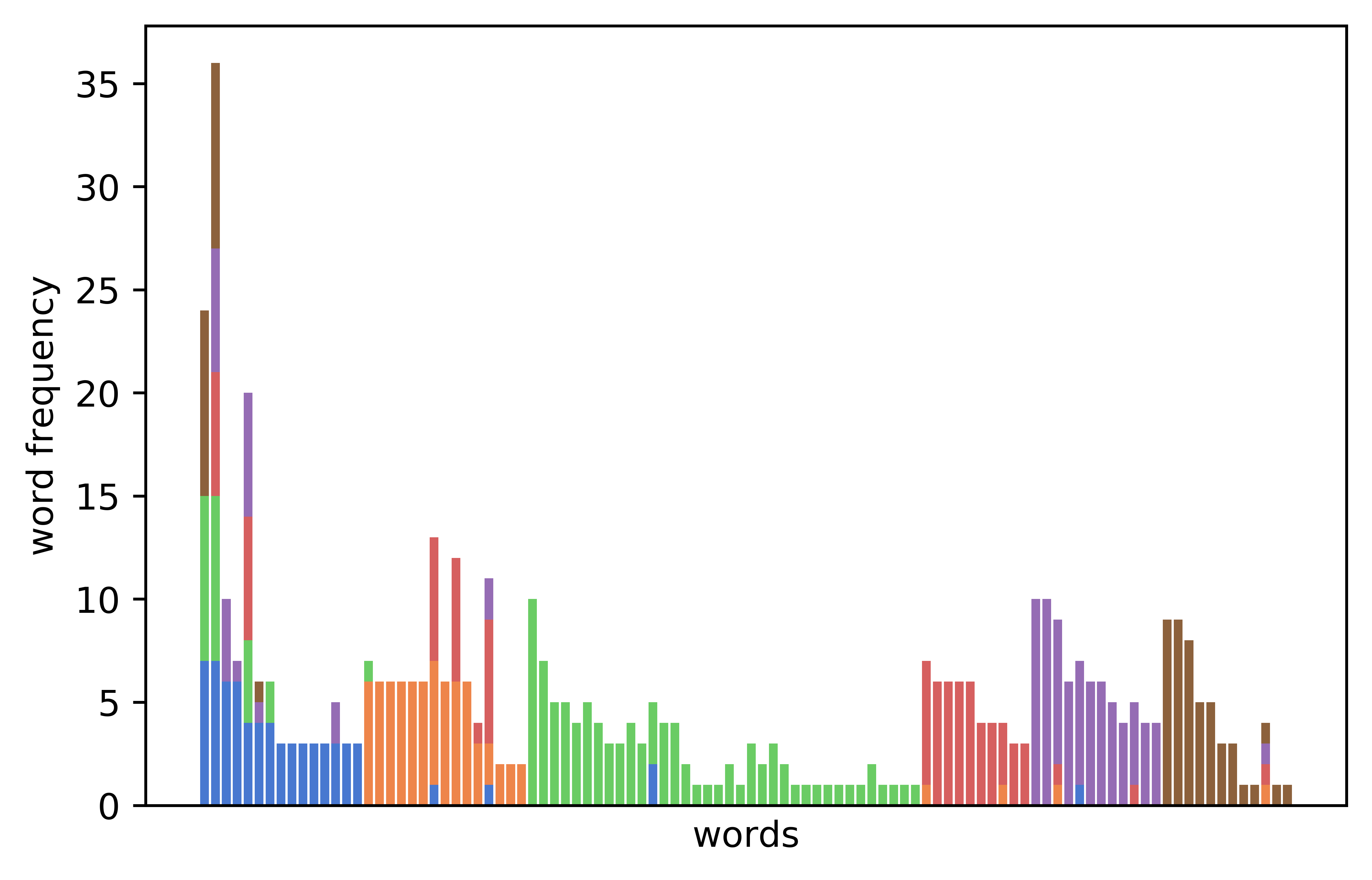}
         \caption{Ground-truth}
         \label{fig:truth}
     \end{subfigure}
     \hfill
     \begin{subfigure}[b]{0.23\textwidth}
         \centering
         \includegraphics[width=\textwidth]{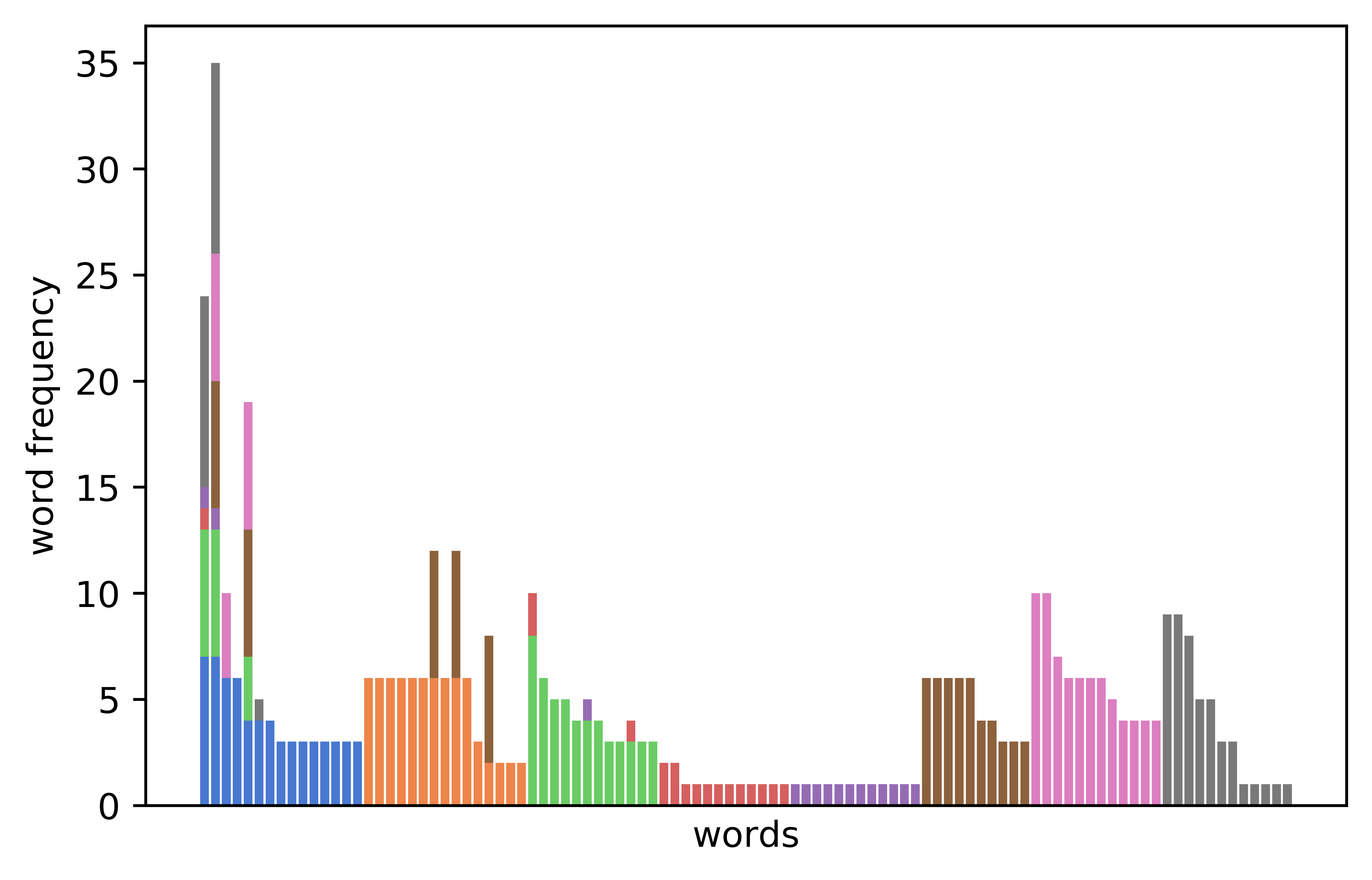}
         \caption{Predicted}
         \label{fig:pred}
     \end{subfigure}
     \hfill
     \begin{subfigure}[b]{0.3\textwidth}
         \centering
         \includegraphics[width=\textwidth]{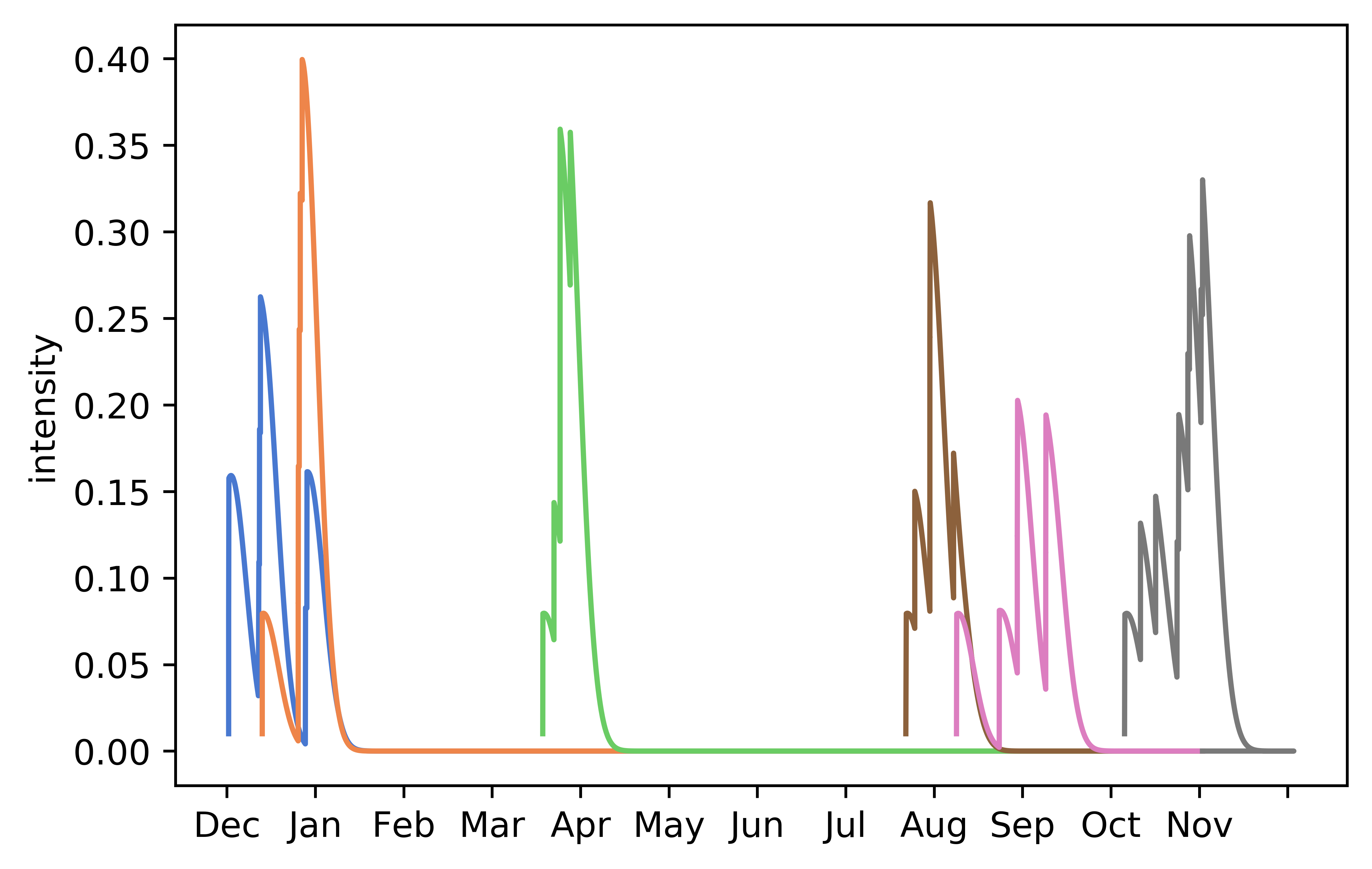}
         \caption{Intensity}
         \label{fig:intensity}
     \end{subfigure}
     \caption{Word and intensity distributions of the transaction sequence from Wall Street Market. Each color indicates a hidden user detected by the proposed approach.}
     \label{fig:illu_word}
\end{figure}

{\bf \noindent Visualization.}
We further show the visualization results on the transaction sequences from Wall Street Market to illustrate the effectiveness of the proposed approach. We investigate our approach for hidden buyer identification from \textit{content} and \textit{temporal dynamics} aspects. To show the content information, we select the top 15 words from each predicted hidden buyer (cluster) and compare the word distributions between the ground-truth and predicted hidden buyers. Figures \ref{fig:truth} and \ref{fig:pred} show the word distribution of the sequence from Wall Street. Each color indicates the word distribution of one anonymized ID, while each bar indicates one word. We can observe that word distributions of predicted hidden buyers are very close to those of ground-truth, which indicates the importance of adopting content information for hidden buyer identification.

To show the information of temporal dynamics, we plot the intensity values of six identified hidden buyers ($\lambda_h$) over time. For the other two identified hidden buyers, since each of them only has one transaction, we omit the intensity curves of them for simplicity. We can observe that these six hidden buyers are active at different months. Meanwhile, due to the self-excitation property of Hawkes process, once a transaction occurs, the intensity increases. Hence, when a hidden buyer becomes active, the following transactions have high chance to be from the same hidden buyer based on Equation \ref{eq:lambda_prior}.

\subsection{Experiments on Transaction Sequences without Ground-truth}
{\bf \noindent Experimental setup.} In this experiment, we apply our algorithm on the transaction sequences without ground-truth from various anonymized IDs. For each darknet market, we select anonymized IDs with at least 50 transactions. We then have 28, 16, and 579 anonymized IDs for Wall Street Market, Empire Market, and Dream Market.

We adopt the \textit{Silhouette coefficient} (Silhouette) and the \textit{topic coherence} ($C_v$) to measure the consistency of clustering results \cite{rousseeuw1987silhouettes,Roder:2015}. Both of these metrics evaluate the clustering performance without ground-truth. Originally, topic coherence evaluates topic models via top-k topic words. In this work, we extract the top-k frequent words from each cluster and evaluate their coherence. If the transactions in a cluster have high coherence in product titles and comments, we can then reasonably consider the cluster of transactions is conducted by one hidden buyer. The metric of topic coherence is implemented by Gensim~\footnote{\url{https://radimrehurek.com/gensim/index.html}}. For Silhouette coefficient, we use the word distribution as the feature vector for each transaction. We report the mean value of each metric over various anonymized IDs in each market.

\begin{table}[]
\centering
\caption{Results of hidden buyer identification on transaction sequences without ground-truth}
\label{tb:cluster_real}
\resizebox{0.4\textwidth}{!}{%
\centering
\begin{tabular}{|c|c|c|c|}
\hline
                              & Approaches   & $C_v$   & Silhouette \\ \hline
\multirow{3}{*}{Wall Street Market}  & HDP          & 0.4941 & -0.0940    \\ \cline{2-4}
                              & DHP          & 0.7361 & -0.0040    \\ \cline{2-4}
                              & Our approach & 0.7668 & 0.0063     \\ \hline
\multirow{3}{*}{Empire Market}         & HDP          & 0.4749 & -0.0784    \\ \cline{2-4}
                              & DHP          & 0.6659 & -0.0154    \\ \cline{2-4}
                              & Our approach & 0.6726 & 0.0026     \\ \hline
\multirow{3}{*}{Dream Market} & HDP          & 0.5367 & 0.1101     \\ \cline{2-4}
                              & DHP          & 0.6667 & 0.1202     \\ \cline{2-4}
                              & Our approach & 0.6735 & 0.1439     \\ \hline
\end{tabular}
}
\end{table}

{\bf \noindent Experimental results.}
As shown in Table \ref{tb:cluster_real}, our proposed approach achieves the best performance in terms of topic coherence and Silhouette coefficient. Specifically, Compared with DHP that does not adopt the vendor information, our approach achieves higher values in $C_v$ and Silhouette coefficient, which indicates the useful of incorporating vendor information for hidden buyer identification. HDP has the lowest values in $C_v$ and Silhouette coefficient among three approaches over three datasets. This could be because HDP does not use the temporal dynamics information. Since the intensity of the transaction is critical for hidden buyer identification, without using the temporal information could lead to poor performance.

\begin{figure}
    \centering
    \includegraphics[width=.45\textwidth]{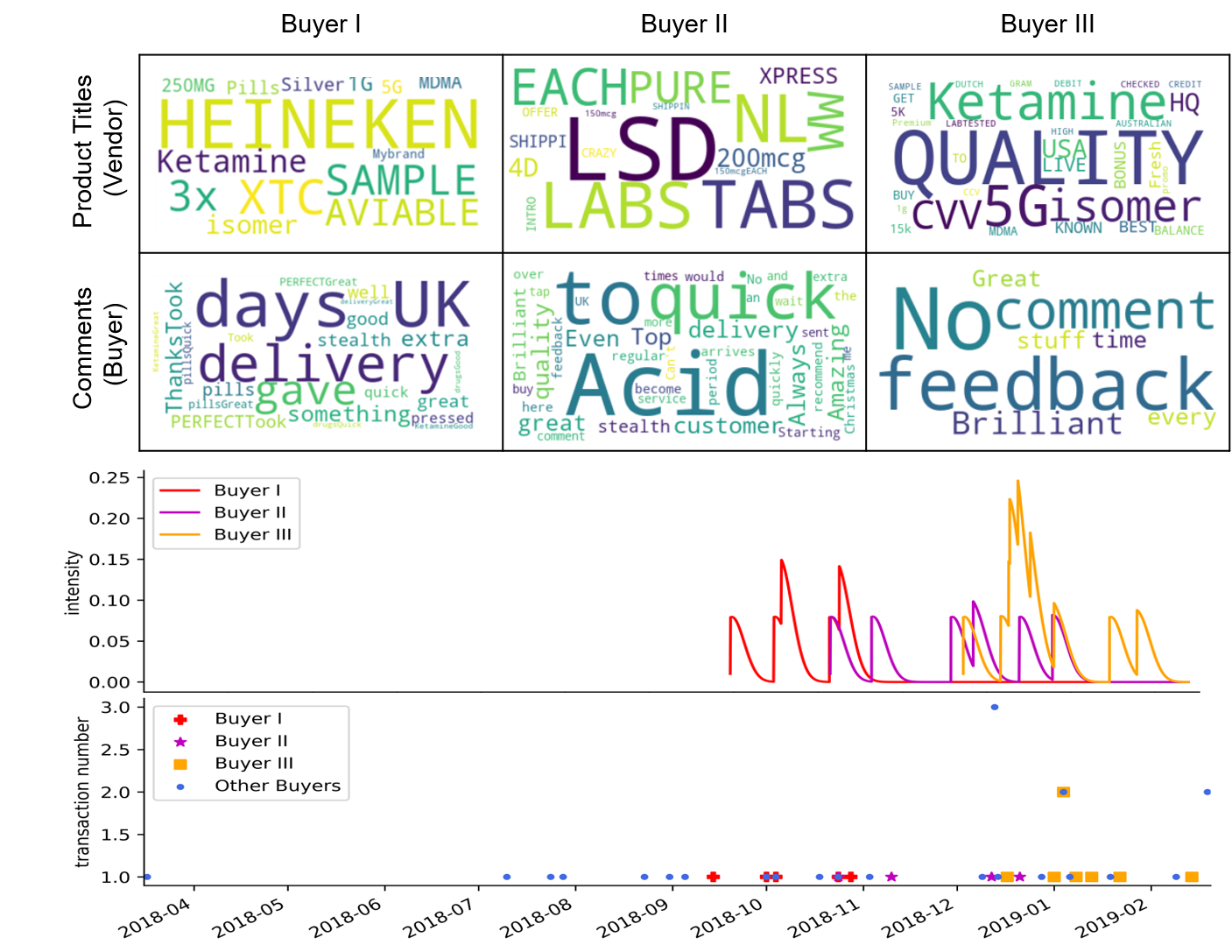}
    \caption{The hidden buyers identified from an anonymized ID ``s**y'' in Empire Market. The first and second rows show the frequent words in product titles and comments associated with each hidden buyer, respectively. The third row show the intensity of each hidden buyer, while the bottom row shows the distribution of transactions over time.}
    \label{fig:cluster_result}
\end{figure}
{\bf \noindent Case study.}
Figure \ref{fig:cluster_result} shows an instance of the hidden buyer identification.
Given a transaction sequence with 94 transactions from an anonymized ID ``s**y'' in Empire Market, we detect 22 hidden buyers by our proposed approach. We show the top 3 hidden buyers who have the highest transaction numbers, i.e., 21, 13, 13. The first and second rows show the top words in product titles and comments from these 3 hidden buyers, respectively. The third row shows the intensity values of the 3 hidden buyers. The last row shows the distribution of transactions from ``s**y'' over time.
We can observe that the transactions from these 3 hidden buyers roughly spread over time. In particular, the time ranges of transactions from Buyer I, II and III are 2018-09-14 to 2018-11-03, 2018-10-24 to 2019-01-22, and 2018-12-13 to 2019-02-14, respectively. The intensities of the three identified hidden buyers also lie in these areas.
Meanwhile, the products bought by the three buyers are different. For example, Buyer I buys products with the frequent word ``HEINEKEN'' in titles, while Buyer II and III buy products with frequent words ``LSD'', and ``Ketamine'', respectively. Although ``Ketamine'' appears in product titles bought by both Buyer I and III, Buyer I and Buyer III have  different comment styles, i.e., Buyer I prefers to write detail comments while Buyer III usually does not comment on the products. Moreover, from the time aspect, Buyer I and III are active in different months. Overall, we can notice that the three hidden buyers detected from the anonymized ID ``s**y'' have different styles in time, product or comment perspective.

\section{Conclusions}
In this paper, we have proposed UNMIX for hidden buyer identification in darknet markets. Due to the unfixed number of hidden buyers, UNMIX adopts the Dirichlet process to group the transactions from one hidden buyer into a cluster. In order to capture the hidden behavior of different buyers, UNMIX uses the Hawkes process to model the transaction time information, the multinomial distribution to model the text information in product titles and comments, and the categorical distribution to model the vendors involved in the transactions.
Experimental results in three real-world darknet markets show that UNMIX achieves the best performance for hidden buyer identification. The case studies also indicate that different hidden buyers identified by UNMIX have different behavior. In the future, we plan to study how to incorporate buyer ratings into our current framework to further improve the performance of hidden buyer identification. Meanwhile, we plan to also investigate linking hidden buyers across different darknet markets.

\section*{Acknowledgments}
This work was supported in part by NSF (1564250, 1937010) and the Department of Energy (DE-OE0000779).


\end{document}